\documentclass{article}

\PassOptionsToPackage{numbers}{natbib}
\usepackage[preprint]{neurips_2026}
\makeatletter
\renewcommand{\@notice}{}
\makeatother

\usepackage[utf8]{inputenc}
\usepackage[T1]{fontenc}
\usepackage{hyperref}
\hypersetup{
  pdftitle={VisualWorldBench: A Fine-Grained Multi-Task Benchmark for Evaluating Visual World Knowledge in MLLMs},
  pdfauthor={Chen Chen, Zheming Liang, Jiaqi Wang},
  pdfkeywords={multimodal large language models, fine-grained recognition, visual world knowledge, benchmark},
  hidelinks
}
\usepackage{url}
\usepackage{booktabs}
\usepackage{amsfonts}
\usepackage{amsmath}
\usepackage{nicefrac}
\usepackage{microtype}
\usepackage{xcolor}
\usepackage{colortbl}
\usepackage{graphicx}

\definecolor{hlrow}{HTML}{EAF4FB}        %
\definecolor{deltacl}{HTML}{2E7D32}      %
\newcommand{\vg}[2]{\textbf{#1}{\tiny\textcolor{deltacl}{\,+#2}}}
\usepackage{subcaption}
\usepackage{multirow}
\usepackage{pifont}      %
\usepackage{enumitem}    %
\usepackage{wrapfig}     %
\usepackage[most]{tcolorbox}  %

\newcommand{\bench}{VisualWorldBench}
\newcommand{\trainsuite}{VisualWorldBench-50K}

\title{VisualWorldBench: A Fine-Grained Multi-Task \\ Benchmark for Evaluating Visual World Knowledge \\ in MLLMs}

\author{%
  Chen Chen$^{1,2}$ \qquad
  Zheming Liang$^{1,2}$ \qquad
  Jiaqi Wang$^{2,3}$\thanks{Corresponding author.} \\
  $^{1}$University of Science and Technology of China \\
  $^{2}$Shanghai Innovation Institute \\
  $^{3}$JD.com
}

\begin{document}

\maketitle

\begin{abstract}
Visual world knowledge is a core capability of multimodal large language models (MLLMs), underpinning knowledge-intensive visual question answering and fine-grained visual reasoning. However, existing benchmarks remain limited in two key respects. First, their category coverage is narrow: most benchmarks span only a small number of categories across a few domains, falling far short of the open-ended scale of real-world visual entities. Second, their task construction is often insufficiently rigorous: models can answer correctly using coarse-grained cues, vague associations, or elimination strategies without truly mastering precise concepts, making it difficult to accurately assess their knowledge of fine-grained entities.
We introduce \bench{}\footnote{Evaluation dataset: \url{https://huggingface.co/datasets/orange0318/VisualWorldBench}. Evaluation code: \url{https://github.com/chenchen0103/VisualWorldBench}.}, a taxonomy-grounded multi-task benchmark built on a hierarchical taxonomy of 21{,}291 leaf-node classes across 8 visual domains, with 32{,}541 evaluation instances whose ground-truth labels cover 15{,}479 categories across four complementary tasks: closed-set recognition, contrastive selection, precise localization, and open-world naming,  covering different directions of knowledge retrieval and varying degrees of answer openness to comprehensively evaluate visual world knowledge.
Most questions are constructed from taxonomically proximate categories, forcing precise discrimination among fine-grained entities. %
Experiments reveal that the strongest closed-source model achieves only 80.4\% average accuracy, exposing capability gaps hidden under conventional benchmarks.
We further construct \trainsuite{}, a training resource covering 50K+ categories. Fine-tuning on \trainsuite{} yields over 21 percentage points improvement on \bench{}, with consistent transfer to public fine-grained recognition and visual knowledge benchmarks.

\end{abstract}

\begin{figure}[h]
\centering
\includegraphics[width=\textwidth]{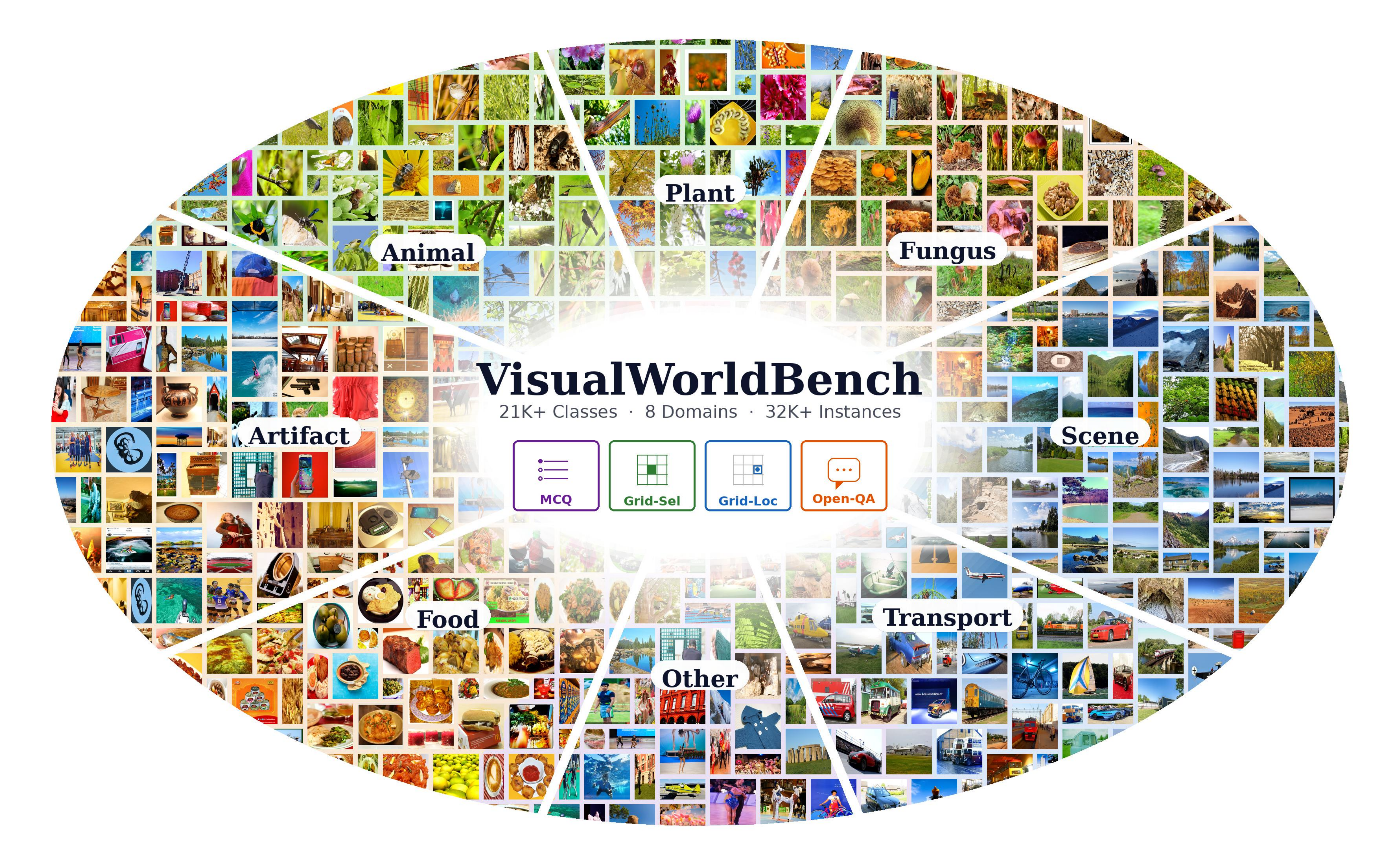}
\caption{Overview of \bench{}. The benchmark spans \textbf{8 visual domains}---Animal, Plant, Fungus, Food, Scene, Transport, Artifact, and Other---with a taxonomy of \textbf{21,291 leaf-node classes} and \textbf{32,541 evaluation instances} covering \textbf{15,479 ground-truth categories}. The benchmark evaluates four complementary tasks: closed-set recognition (MCQ), contrastive selection (Grid-Sel), spatial localization (Grid-Loc), and open-world naming (Open-QA).}

\label{fig:teaser}
\end{figure}

\section{Introduction}
\label{sec:introduction}

The visual world comprises millions of fine-grained entities---species, cultivars, architectural styles, vehicle models. Yet whether current multimodal large language models (MLLMs)~\cite{gpt4o,qwen2vl}, driven by large-scale vision-language pretraining~\cite{clip,siglip2}, can truly discriminate among such entities rather than merely categorize them at a coarse level remains an open question. 
It can be regarded that fine-grained visual entity concepts are a form of \textit{visual world knowledge}. Whether a model maintains clear discrimination boundaries among visually similar entities reflects the extent of visual world knowledge it has internalized, which underpins knowledge-intensive visual question answering and fine-grained visual reasoning.%

However, existing benchmarks fall short along two complementary dimensions.
(1) \textbf{Limited category coverage.} Classic fine-grained benchmarks, such as CUB-200~\cite{cub200}, Stanford Dogs~\cite{stanforddogs}, and FGVC-Aircraft~\cite{aircraft}, focus on narrow domains and limited category coverage. Even larger-scale extensions such as iNat2021~\cite{inat2021} and BIOSCAN-5M~\cite{bioscan5m} remain largely concentrated in biological domains. Broader entity-knowledge benchmarks, including InfoSeek~\cite{infoseek}, SimpleVQA~\cite{simplevqa}, and WorldVQA~\cite{worldvqa}, expand beyond single-domain recognition, yet still cover only a limited set of categories, falling far short of the open-ended scale of real-world visual entities.
(2) \textbf{Insufficient evaluation rigor.} General MLLM benchmarks such as MMBench~\cite{mmbench}, MMMU~\cite{mmmu}, and MMMU-Pro~\cite{mmmupro} evaluate multimodal competence rather than precise entity-level visual knowledge. Traditional fine-grained benchmarks are typically formulated as closed-set classification, which does not align with prompt-based MLLM evaluation. Meanwhile, existing entity-knowledge benchmarks often rely on task formats such as attribute QA, naming, or coarse-grained multiple-choice, where models can succeed through vague associations or elimination strategies without truly distinguishing between closely related fine-grained entities. As a result, current evaluations often fail to verify whether a model has learned precise conceptual boundaries for visual entities. %

To address these limitations, we introduce \bench{}, a taxonomy-grounded multi-task benchmark for evaluating visual world knowledge in MLLMs. Built on a taxonomy of 21{,}291 leaf-node classes across 8 visual domains, it comprises 32{,}541 evaluation instances whose ground-truth labels cover 15{,}479 distinct categories across the four tasks. Table~\ref{tab:bench_comparison} compares \bench{} with representative prior work.

\begin{table}[t]
\centering
\caption{Comparison with existing visual world knowledge and fine-grained recognition benchmarks.
\bench{} is the only benchmark that combines large-scale fine-grained coverage, multi-domain scope, and multiple complementary task formats in a single MLLM-oriented evaluation.
For \bench{}, \#Classes counts distinct ground-truth categories across the four tasks; its underlying taxonomy contains 21{,}291 leaf-node classes.
$^{*}$For InfoSeek-Human, 806 denotes the number of visual entities referenced by the questions, not distinct answer categories.
\textit{Tax.}: hierarchical taxonomy provided.
\textit{FG}: fine-grained.
}
\label{tab:bench_comparison}
\small
\setlength{\tabcolsep}{4pt}
\resizebox{\textwidth}{!}{%
\begin{tabular}{@{}lccccccc@{}}
\toprule
\textbf{Benchmark} & \textbf{Year} & \textbf{Coverage} & \textbf{\#Classes} & \textbf{\#Eval} & \textbf{Tasks} & \textbf{Tax.} & \textbf{MLLM} \\
\midrule
\multicolumn{8}{@{}l}{\textit{Fine-grained recognition benchmarks}} \\
CUB-200~\cite{cub200}             & 2011 & Bird     & 200    & 5.8K   & Cls            & \ding{55} & \ding{55} \\
Stanford Dogs~\cite{stanforddogs} & 2011 & Dog      & 120    & 8.6K   & Cls            & \ding{55} & \ding{55} \\
FGVC-Aircraft~\cite{aircraft}     & 2013 & Aircraft & 100    & 3.3K   & Cls            & \checkmark & \ding{55} \\
iNat2021~\cite{inat2021}          & 2021 & Bio      & 10K    & 500K   & Cls            & \checkmark & \ding{55} \\
FG-BMK~\cite{fgbmk}              & 2026 & FG       & --     & 1.01M  & QA+Ret+Cls     & \ding{55} & \checkmark \\
\midrule
\multicolumn{8}{@{}l}{\textit{MLLM general and knowledge benchmarks}} \\
MMBench~\cite{mmbench}            & 2024 & General  & --     & 2.4K   & Perception+Reasoning & \ding{55} & \checkmark \\
MMMU-Pro~\cite{mmmupro}           & 2024 & Academic & --     & 5.2K   & MC reasoning         & \ding{55} & \checkmark \\
InfoSeek-Human~\cite{infoseek}    & 2023 & Entity   & 806$^{*}$ & 8.9K   & Knowledge QA   & \ding{55} & \checkmark \\
WorldVQA~\cite{worldvqa}          & 2026 & Vision knowledge & -- & 3.5K & Knowledge QA   & \checkmark & \checkmark \\
SimpleVQA~\cite{simplevqa}        & 2025 & Vision knowledge & -- & 2.0K & Knowledge QA   & \ding{55} & \checkmark \\
\midrule
\bench{} (Ours)                   & 2026 & \textbf{Multi-dom FG} & \textbf{15,479} & \textbf{32K+} & \textbf{MCQ+Sel+Loc+Open} & \checkmark & \checkmark \\
\bottomrule
\end{tabular}%
}
\end{table}

Our benchmark has three distinctive features.
(1)~\textbf{Large-scale category coverage}: its ground-truth labels cover 15{,}479 categories across 8 visual domains, supported by a taxonomy of 21{,}291 leaf-node classes.
(2)~\textbf{Taxonomy-driven precise evaluation}: most questions are constructed from taxonomically proximate categories, requiring models to distinguish precise conceptual boundaries among fine-grained entities and preventing shortcuts based on vague knowledge or elimination strategies.
(3)~\textbf{Comprehensive task coverage}: this principle is instantiated through four complementary tasks, including text-to-image selection, image-to-text selection, target localization, and open-ended naming, covering different directions of knowledge retrieval and varying degrees of answer openness. 

Systematic evaluation across diverse closed-source and open-source models reveals substantial gaps in precise visual world knowledge. Large performance differences persist across task formats even within the same model family: for example, InternVL3.5-8B reaches 35.5\% on closed-set recognition but only 7.4\% on 3$\times$3 localization, and similar patterns appear consistently across the Qwen, InternVL, and Gemma families. These results suggest that different task formats impose different demands on precise entity knowledge, and that closed-set accuracy alone substantially overestimates a model's mastery of visual world knowledge. 
Based on these findings, we further construct \trainsuite{}, a training resource covering 50K+ categories with supervision formats aligned to the benchmark. After fine-tuning, models improve by over 21 percentage points on average on \bench{}, and the gains transfer to multiple public fine-grained recognition benchmarks, including CUB-200, Stanford Dogs, and iNaturalist~2021, as well as the visual entity knowledge benchmark WorldVQA. These results demonstrate that the capability gaps exposed by \bench{} are actionable and can be effectively reduced with targeted entity-level supervision. Our main contributions are as follows:

\begin{itemize}
    \item We introduce \bench{}, a taxonomy-grounded multi-task benchmark for visual world knowledge, comprising 32{,}541 evaluation instances with 15{,}479 ground-truth categories across 8 domains, built on a taxonomy of 21{,}291 leaf-node classes.
    \item We propose a taxonomy-driven evaluation paradigm in which questions are constructed from taxonomically proximate categories and instantiated through four complementary task formats, enabling rigorous assessment of precise fine-grained entity knowledge.
    \item We show that conventional closed-set evaluation substantially overestimates model capability, while more discriminative and open-ended settings expose large gaps in precise visual world knowledge; we further construct \trainsuite{}, a 50K+-category training resource that improves performance on \bench{} by over 21 percentage points and transfers consistently to public benchmarks.
\end{itemize}

\section{Related Work}
\label{sec:related_work}

\paragraph{Fine-grained recognition benchmarks.}
Fine-grained visual recognition has been studied extensively through domain-specific classification datasets spanning birds~\cite{cub200,nabirds}, dogs~\cite{stanforddogs}, cars~\cite{stanfordcars,compcars}, aircraft~\cite{aircraft}, food~\cite{food101}, flowers~\cite{flowers102}, and pets~\cite{oxfordpet}.
These datasets have driven substantial methodological progress, but they are confined to single domains with limited category counts and have become largely saturated under modern vision models.
Subsequent efforts scale category coverage: iNat2021~\cite{inat2021} introduces 10K biological species with a taxonomic hierarchy, and BIOSCAN-5M~\cite{bioscan5m} provides millions of insect specimens for cross-modal alignment.
However, both remain within the biological domain and adopt traditional closed-set protocols rather than prompt-based evaluation.
More recently, FG-BMK~\cite{fgbmk} unifies established fine-grained datasets into MLLM-oriented question-answering and retrieval protocols, but its category coverage remains constrained by the scale of the original datasets.

\paragraph{MLLM evaluation and visual grounding.}
Broad-coverage MLLM benchmarks~\cite{mmbench,mmmu,mmstar} have established comprehensive evaluation protocols for multimodal reasoning, but fine-grained entity recognition constitutes only a small fraction of their test suites and is not diagnosed in isolation.
A separate line of work targets visual entity knowledge directly: InfoSeek~\cite{infoseek} and Enc-VQA~\cite{encvqa} evaluate factual and encyclopedic knowledge about entities through question answering, yet do not assess whether models can discriminate among visually confusable alternatives or localize them spatially.
On the grounding side, referring expression comprehension~\cite{kazemzadeh2014referitgame,mao2016refcocog} and grounded MLLMs~\cite{chen2023shikra,ferret,glamm} require localizing objects described by natural-language expressions; here, the challenge lies in resolving spatial and attribute-based ambiguity among common objects, which differs fundamentally from discriminating taxonomically adjacent fine-grained categories that share near-identical visual appearance.

\section{Benchmark}
\label{sec:benchmark}

The central design principle of \bench{} is that all tasks share a single hierarchical taxonomy.
This structure provides taxonomy-aware distractor sampling for evaluation tasks, category-level references for quality control, and a unified framework for organizing the training resource.
We describe the taxonomy (\S\ref{sec:taxonomy}), task design (\S\ref{sec:task_design}), quality control (\S\ref{sec:qc}), and training resource (\S\ref{sec:dataset}) in turn; overall benchmark statistics are provided in Appendix Table~\ref{tab:bench_stats}.

\begin{figure}[t]
\centering
\includegraphics[width=\textwidth]{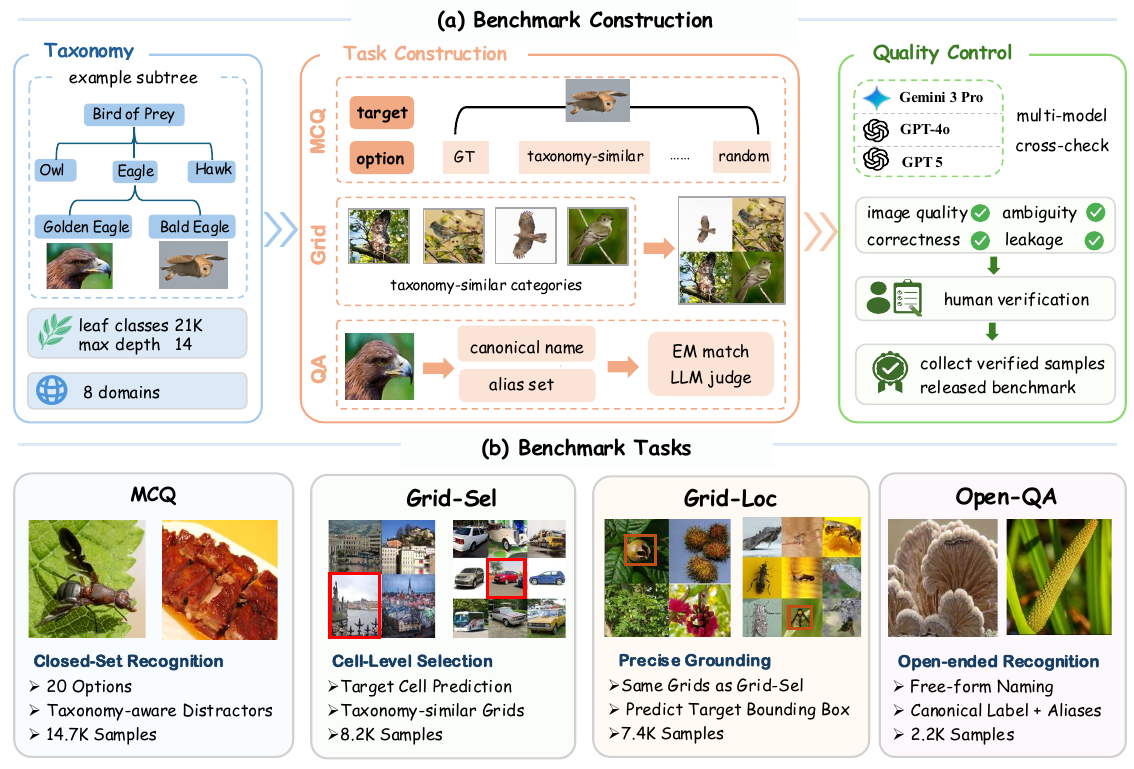}
\caption{Taxonomy-grounded construction and four-format evaluation design of \bench{}.
\textbf{(a)}~Benchmark construction: a hierarchical taxonomy (illustrated with an example subtree), multi-format task generation, and two-stage quality control (model-assisted triage by three frontier models followed by human verification and release balancing).
\textbf{(b)}~The four evaluation formats—MCQ for closed-set recognition, Grid-Sel for contrastive selection, Grid-Loc for precise localization, and Open-QA for open-world naming—with their task setup, sample counts, and design choices.}
\label{fig:benchmark_overview}
\end{figure}

\subsection{Taxonomy and Data Sources}
\label{sec:taxonomy}

\bench{} is built on a hierarchical taxonomy containing 21{,}291 leaf-node classes with a maximum depth of 14 levels (Figure~\ref{fig:benchmark_overview}a), spanning 8 visual domains: animal, plant, fungus, artifact, transport, food, scene, and other.
The taxonomy is derived from the Bamboo category tree~\cite{bamboo} and further expanded with supplementary public sources to improve category coverage across domains.
To make the taxonomy suitable for benchmark construction, we further curate both the source category set and the associated image pool through normalization, deduplication, filtering, and quality control.
After additional image-quality filtering that removes low-quality images (too dark, too small, blurry, low saturation, etc.), the released benchmark covers 15{,}479 ground-truth categories across all tasks while nearly all taxonomy classes appear as distractors in MCQ or Grid tasks.
Details of taxonomy construction, including per-domain class distributions and depth statistics, are provided in the appendix.

\subsection{Task Design}
\label{sec:task_design}

\begin{figure}[t]
\centering
\includegraphics[width=\textwidth]{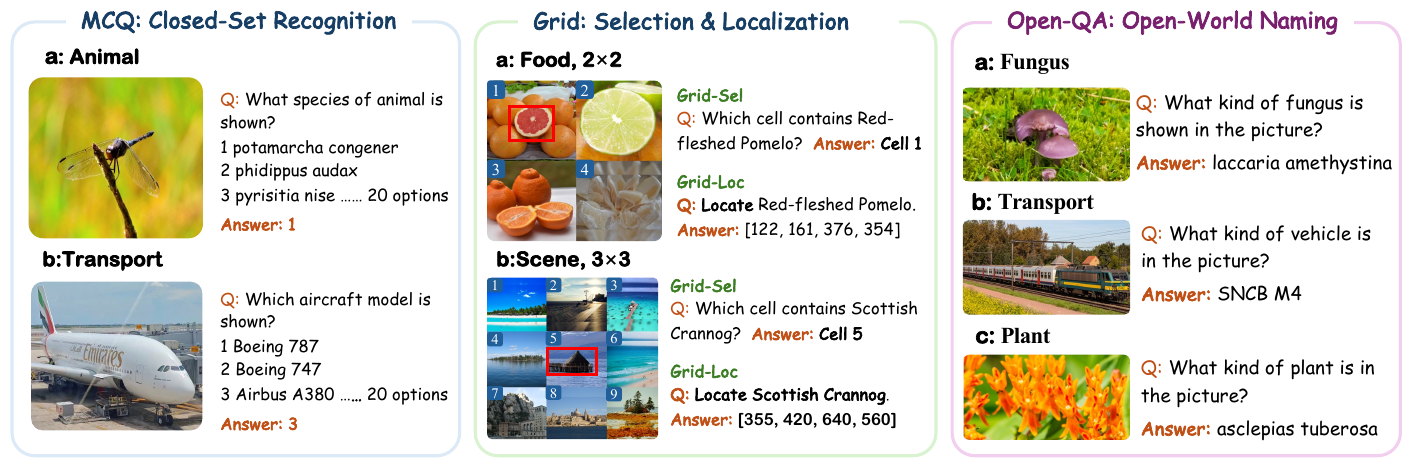}
\caption{Examples of the four \bench{} tasks, all grounded in a shared 21{,}291-class taxonomy.
\textbf{MCQ} (closed-set recognition): the model selects the correct category from 20 taxonomy-aware options (sibling, near, and random distractors).
\textbf{Grid-Sel / Grid-Loc} (contrastive selection and precise localization): the model identifies or localizes the target category among visually similar alternatives in a composite grid. Grid-Sel requires selecting the correct cell, while Grid-Loc requires outputting a precise bounding box on the same grid layout.
\textbf{Open-QA} (open-world naming): the model generates the fine-grained category name without any candidate set.}
\label{fig:task_examples}
\end{figure}

We design four task formats to probe four core capabilities: discriminating the correct category from confusable alternatives (MCQ), selecting the target image among visually similar alternatives (Grid-Sel), precisely localizing it with a bounding box (Grid-Loc), and producing its precise name without any candidate set (Open-QA).
Figure~\ref{fig:task_examples} illustrates representative examples of each task.

\paragraph{MCQ: closed-set recognition.}
Each sample presents an image together with 20 candidate class names, and the model must select the single correct category.
We use 20 options, substantially more than the typical 4--5 in general-purpose benchmarks, to reduce the effect of random guessing.
Unlike standard evaluations that rely on random distractors, we construct options using hierarchical distance in the taxonomy: \textit{sibling} categories share the same parent node, \textit{near} categories are taxonomically close but not siblings, and \textit{random} categories are sampled from farther away in the hierarchy.
Approximately 60\% of distractors come from taxonomy neighbors (sibling + near), requiring fine-grained discrimination among semantically and visually related candidates (Figure~\ref{fig:benchmark_overview}b).

\paragraph{Grid-Sel / Grid-Loc: contrastive selection and precise localization.}
Grid-Sel and Grid-Loc evaluate whether a model can identify a target category when multiple visually similar alternatives are presented simultaneously in a controlled composite-image setting.
Each sample is formed by tiling the target image together with images from taxonomically similar categories into a 2$\times$2 or 3$\times$3 grid, with one category per cell; more than 99\% of distractor cells are populated by \textit{sibling} categories of the target.
Grid-Sel requires the model to identify the target cell, measuring contrastive selection ability under tightly controlled visual competition; Grid-Loc uses the same grid layouts but requires the model to output a bounding box around the target object, measuring precise localization under the same controlled setup.
The 3$\times$3 setting is more demanding due to smaller cell images and more confusable alternatives.

\paragraph{Open-QA: open-world naming.}
Given an image and a natural-language question (e.g., ``What species of bird is this?''), the model must generate the fine-grained category name directly, with no candidate options provided.

\paragraph{Evaluation metrics.}
MCQ and Grid-Sel are evaluated by accuracy.
Grid-Loc is evaluated by IoU@0.5 after applying a fixed bounding-box extraction and normalization procedure. Parsing and failure-handling details are provided in the appendix.
Open-QA is evaluated by EM+LLM Judge: we first match the extracted answer against the canonical label and alias set, then use a judge model to adjudicate semantically equivalent non-exact matches; full judging details are provided in the appendix.

\subsection{Quality Control}
\label{sec:qc}

We apply a two-stage release pipeline to all tasks: \textit{annotation verification} followed by \textit{release balancing and filtering}.

\paragraph{Annotation verification.}
We use three frontier closed-source models as \emph{triage signals}: when at least two models produce an incorrect answer, the sample is flagged for human review.
Final keep/remove decisions follow predefined annotation criteria (label correctness, visual identifiability, ambiguity, and leakage checks) under human review; flagged samples are never removed based on model behavior alone.
We additionally conduct a random audit on non-flagged samples to ensure model-assisted triage does not miss systematic issues; details are reported in Appendix~\ref{sec:qc_audit}.

\paragraph{Release balancing and filtering.}
The final evaluation set is assembled from the verified pool under category-coverage, domain-representation, distractor-structure, and image-quality constraints, subject to a \textit{class-safe constraint} that no category may be lost from the union of all four tasks.
Per-stage statistics, audit protocols, and detailed task-specific release rules are provided in the appendix.

\subsection{\trainsuite{}}
\label{sec:dataset}

In addition to the benchmark, we construct \trainsuite{}, a broader fine-grained training resource covering over 50K categories.
It is anchored by large-scale closed-set MCQ supervision and uses taxonomy-aware near-neighbor distractors to strengthen fine-grained category discrimination.
On top of this MCQ backbone, we introduce smaller amounts of grounding and open-ended supervision as complementary signals, exposing models to spatial alignment and category naming beyond closed-set recognition alone.
In total, \trainsuite{} comprises approximately 7.7M training instances collected under the benchmark taxonomy from publicly available fine-grained datasets and other public image sources, with MCQ accounting for roughly 90\% of records and grounding/QA each contributing approximately 5\%.
As shown in Section~\ref{sec:experiments}, the capability gaps exposed by \bench{} can be effectively reduced through this targeted supervision.
To reduce contamination risk, we exclude Cars-196 test images and iNaturalist~2021 validation images from the training pool, as detailed in Appendix~\ref{sec:data_integrity}.
Its construction details and experimental use are described in Section~\ref{sec:experiments} and Appendix~\ref{sec:training_details}.

\section{Experiments}
\label{sec:experiments}

\subsection{Experimental Setup}
\label{sec:exp_setup}

\paragraph{Models.}
We evaluate three groups of models on \bench{}.
(1)~\textit{Closed-source models}: representative frontier commercial MLLMs from Google, OpenAI, ByteDance, and Xiaomi---Gemini 3.1 Pro~\cite{gemini}, GPT-5.4~\cite{gpt54}, GPT-5.2~\cite{gpt52}, Doubao-Seed 2.0 Pro~\cite{doubao_seed}, and Mimo-V2-Omni~\cite{mimo_v2_omni}.
(2)~\textit{Open-source models}: spanning four model families (Kimi, Qwen, InternVL, Gemma) across multiple parameter scales to reduce dependence on any single architecture: Kimi-K2.5~\cite{team2026kimi}, Qwen3-VL-2B/4B/8B~\cite{qwen3vl}, Qwen3.5-4B/9B/27B~\cite{qwen3_5}, InternVL3.5-4B/8B~\cite{internvl3_5}, Gemma-4-31B, and Gemma-4-26B-A4B (MoE)~\cite{gemma4}.
(3)~\textit{SFT models}: Qwen3-VL-8B/4B/2B and InternVL3.5-8B fine-tuned on \trainsuite{} (Section~\ref{sec:dataset}) to test whether the capability gaps exposed by \bench{} can be reduced through targeted fine-grained supervision.
Throughout the paper, Qwen3-VL-2B/4B/8B and InternVL3.5-8B rows without $^\dagger$ denote the original instruct models, while rows with $^\dagger$ denote the corresponding \trainsuite{}-tuned variants.

\paragraph{Training data.}
All SFT models are fine-tuned on \trainsuite{} described in Section~\ref{sec:dataset}, a broader fine-grained training resource covering over 50K classes and anchored by large-scale MCQ supervision with taxonomy-aware near-neighbor distractors.
Smaller grounding and open-ended subsets provide complementary signals beyond closed-set discrimination alone.
Training implementation details and dataset composition are provided in Appendix~\ref{sec:training_details}.

\paragraph{Evaluation protocol.}
Evaluation metrics follow the definitions in Section~\ref{sec:benchmark}.
Grid-Sel and Grid-Loc are evaluated at both 2$\times$2 and 3$\times$3 resolutions; we report both settings as well as their within-task average.
The reported overall average (\textit{Avg}) is the arithmetic mean of MCQ, Sel Avg, Loc Avg, and Open-QA (EM+LLM Judge).
Unless otherwise stated, all models are evaluated in a zero-shot setting with greedy decoding (temperature$=$0); images are fed at each model's native input resolution.
Additional training and evaluation details are provided in the appendix.

\subsection{Main Results}
\label{sec:main_results}

Table~\ref{tab:main_results} presents the main results on \bench{} across closed-source models, open-source models, and our best SFT models.

\begin{table}[t]
\centering
\caption{Main results on \bench{}.
\textit{MCQ}: MCQ accuracy;
\textit{Sel}: Grid-Sel accuracy (2$\times$2 / 3$\times$3 / avg);
\textit{Loc}: Grid-Loc IoU@0.5 (2$\times$2 / 3$\times$3 / avg);
\textit{Open}: Open-QA accuracy (EM+LLM Judge);
\textit{Avg}: arithmetic mean of MCQ, Sel Avg, Loc Avg, and Open-QA.
$^\dagger$Trained on \trainsuite{} (Section~\ref{sec:dataset}).
Best result per column in \textbf{bold}.
}
\label{tab:main_results}
\setlength{\tabcolsep}{3pt}
\resizebox{\textwidth}{!}{%
\begin{tabular}{l | c ccc ccc c c}
\toprule
& & \multicolumn{3}{c}{Sel} & \multicolumn{3}{c}{Loc} & & \\
\cmidrule(lr){3-5} \cmidrule(lr){6-8}
Model & MCQ & 2$\times$2 & 3$\times$3 & Avg & 2$\times$2 & 3$\times$3 & Avg & Open & Avg \\
\midrule
\multicolumn{10}{@{}l}{\textit{Closed-source models}} \\
Gemini 3.1 Pro
  & \textbf{86.0} & \textbf{94.2} & \textbf{87.3} & \textbf{90.8} & \textbf{84.6} & \textbf{83.3} & \textbf{84.0} & 60.6 & \textbf{80.4} \\
Doubao-Seed 2.0 Pro
  & 75.1 & 77.9 & 71.6 & 74.8 & 71.4 & 69.7 & 70.6 & \textbf{74.5} & 73.8 \\
Mimo-V2-Omni
  & 65.5 & 76.7 & 68.7 & 72.7 & 47.3 & 44.2 & 45.8 & 40.0 & 56.0 \\
GPT-5.4
  & 67.9 & 76.4 & 65.8 & 71.1 & 35.1 & 22.0 & 28.6 & 37.4 & 51.2 \\
GPT-5.2
  & 65.5 & 72.7 & 62.6 & 67.7 & 25.6 & 20.2 & 22.9 & 35.0 & 47.8 \\
\midrule
\multicolumn{10}{@{}l}{\textit{Open-source models}} \\
Kimi-K2.5
  & 68.6 & 77.7 & 71.5 & 74.6 & 71.4 & 50.8 & 61.1 & 47.2 & 62.9 \\
Qwen3.5-27B
  & 59.1 & 68.1 & 63.9 & 66.0 & 60.4 & 56.2 & 58.3 & 28.7 & 53.0 \\
Gemma-4-31B
  & 60.5 & 65.3 & 55.6 & 60.5 & 46.5 & 36.1 & 41.3 & 27.2 & 47.4 \\
Qwen3.5-9B
  & 52.4 & 60.8 & 53.9 & 57.4 & 54.6 & 45.2 & 49.9 & 26.8 & 46.6 \\
Qwen3-VL-8B
  & 52.0 & 47.7 & 45.3 & 46.5 & 38.7 & 32.5 & 35.6 & 29.1 & 40.8 \\
Gemma-4-26B-A4B
  & 56.0 & 57.4 & 46.6 & 52.0 & 16.6 & 10.9 & 13.8 & 24.0 & 36.5 \\
Qwen3.5-4B
  & 44.1 & 49.9 & 33.6 & 41.8 & 43.6 & 37.2 & 40.4 & 19.6 & 36.5 \\
Qwen3-VL-4B
  & 47.8 & 40.9 & 35.1 & 38.0 & 28.8 & 24.0 & 26.4 & 22.5 & 33.7 \\
InternVL3.5-8B
  & 35.5 & 44.4 & 31.6 & 38.0 & 16.7 & 7.4 & 12.1 & 10.3 & 24.0 \\
InternVL3.5-4B
  & 33.1 & 38.7 & 29.3 & 34.0 & 9.2 & 8.2 & 8.7 & 7.1 & 20.7 \\
Qwen3-VL-2B
  & 28.0 & 15.6 & 10.3 & 13.0 & 28.9 & 19.5 & 24.2 & 15.8 & 20.2 \\
\midrule
\multicolumn{10}{@{}l}{\textit{SFT models}} \\
\rowcolor{hlrow}
Qwen3-VL-2B$^\dagger$
  & \vg{70.7}{42.7} & \vg{46.5}{30.9} & \vg{31.3}{21.0} & \vg{38.9}{25.9} & \vg{54.3}{25.4} & \vg{50.8}{31.3} & \vg{52.6}{28.4} & \vg{18.5}{2.7} & \vg{45.2}{25.0} \\
\rowcolor{hlrow}
Qwen3-VL-4B$^\dagger$
  & \vg{76.8}{29.0} & \vg{74.6}{33.7} & \vg{61.3}{26.2} & \vg{68.0}{30.0} & \vg{64.4}{35.6} & \vg{60.5}{36.5} & \vg{62.5}{36.1} & \vg{26.5}{4.0} & \vg{58.5}{24.8} \\
\rowcolor{hlrow}
Qwen3-VL-8B$^\dagger$
  & \vg{80.0}{28.0} & \vg{76.0}{28.3} & \vg{70.5}{25.2} & \vg{73.3}{26.8} & \vg{62.8}{24.1} & \vg{64.3}{31.8} & \vg{63.6}{28.0} & \vg{33.9}{4.8} & \vg{62.7}{21.9} \\
\rowcolor{hlrow}
InternVL3.5-8B$^\dagger$
  & \vg{69.3}{33.8} & \vg{69.4}{25.0} & \vg{59.4}{27.8} & \vg{64.4}{26.4} & \vg{53.0}{36.3} & \vg{37.9}{30.5} & \vg{45.4}{33.4} & \vg{14.1}{3.8} & \vg{48.3}{24.3} \\
\bottomrule
\end{tabular}%
}
\end{table}

\paragraph{Visual world knowledge is multi-dimensional, with strong cross-task dissociations.}
Performance varies dramatically across tasks: even models with reasonable MCQ accuracy can fail sharply on Grid-Loc.
For example, InternVL3.5-8B, InternVL3.5-4B, and Gemma-4-26B-A4B achieve only 7.4\%, 8.2\%, and 10.9\% on Loc-3$\times$3, despite MCQ scores of 35.5\%, 33.1\%, and 56.0\%, respectively.
This indicates that precise localization is a systematic bottleneck for current MLLMs, and that closed-set accuracy alone substantially overestimates fine-grained capability.
The performance hierarchy is broadly consistent across the Qwen, InternVL, and Gemma families, suggesting that \bench{} captures general capability differences rather than architecture-specific artifacts.
Moreover, even the best closed-source model (Gemini~3.1~Pro) reaches only 80.4\% overall average, while most open-source models remain below 53\%, confirming that the benchmark is far from saturated.

\paragraph{The exposed capability gaps are actionable under targeted supervision.}
Models fine-tuned on \trainsuite{} achieve substantial gains, particularly on the dimensions \bench{} identifies as weak.
The most striking case appears at the smallest scale: Qwen3-VL-2B$^\dagger$ reaches 70.7\% on MCQ and 45.2\% on Avg, improving over its untuned counterpart by 42.7 and 25.0 percentage points and surpassing the untuned Qwen3-VL-8B instruct model (Avg 40.8\%).
Gains are consistent across all three Qwen3-VL scales and InternVL3.5-8B on MCQ, Grid-Sel, and Grid-Loc alike, confirming that the gaps are not merely diagnostic but can be effectively reduced through targeted fine-grained supervision.

\subsection{Generalization to Public Fine-Grained Benchmarks}
\label{sec:transfer}

To assess whether gains from \trainsuite{} generalize beyond \bench{}, we evaluate the trained models on five widely used public fine-grained MCQ benchmarks (CUB-200, Stanford Dogs, Aircraft, Flowers, Pets), iNaturalist 2021, and WorldVQA.

\begin{table}[t]
\centering
\caption{Generalization to widely used public fine-grained benchmarks.
Bird--Pet: 4-option MCQ accuracy on five public fine-grained benchmarks.
\textit{iNat21}: leaf-level classification accuracy on the animal and plant splits of iNaturalist 2021, plus their average.
\textit{WorldVQA}: open-ended visual QA accuracy (LLM-judged, structured answer format).
$^\dagger$Trained on \trainsuite{}.
}
\label{tab:transfer}
\setlength{\tabcolsep}{2.5pt}
\resizebox{\textwidth}{!}{%
\begin{tabular}{l | cccccc | ccc | c}
\toprule
& \multicolumn{6}{c|}{Public MCQ} & \multicolumn{3}{c|}{iNat21} & \\
\cmidrule(lr){2-7} \cmidrule(lr){8-10}
Model
  & Bird & Dog & Air. & Flwr. & Pet
  & Avg
  & Animal & Plant & Avg
  & WorldVQA \\
\midrule
Qwen3-VL-2B
  & 49.6 & 57.6 & 55.0 & 75.4 & 79.1 & 63.3 & 27.3 & 30.6 & 29.0 & 8.2 \\
Qwen3-VL-4B
  & 55.2 & 74.0 & 57.1 & 85.9 & 93.0 & 73.1 & 35.9 & 36.1 & 36.0 & 10.2 \\
Qwen3-VL-8B
  & 59.3 & 71.5 & 60.7 & 86.8 & 88.1 & 73.3 & 40.2 & 41.3 & 40.8 & 11.2 \\
InternVL3.5-8B
  & 41.7 & 55.4 & 42.2 & 54.1 & 79.8 & 54.6 & 63.0 & 50.3 & 56.7 & 4.8 \\
\midrule
\rowcolor{hlrow}
Qwen3-VL-2B$^\dagger$
  & \vg{62.7}{13.1} & \vg{75.4}{17.8} & \vg{73.5}{18.5} & \vg{89.3}{13.9} & \vg{88.1}{9.0} & \vg{77.8}{14.5} & \vg{51.5}{24.2} & \vg{38.9}{8.3} & \vg{45.2}{16.2} & \vg{13.4}{5.2} \\
\rowcolor{hlrow}
Qwen3-VL-4B$^\dagger$
  & \vg{72.1}{16.9} & \vg{75.8}{1.8} & \vg{75.8}{18.7} & \vg{90.1}{4.2} & \vg{94.0}{1.0} & \vg{81.6}{8.5} & \vg{47.1}{11.2} & \vg{40.0}{3.9} & \vg{43.6}{7.6} & \vg{14.9}{4.7} \\
\rowcolor{hlrow}
Qwen3-VL-8B$^\dagger$
  & \vg{76.3}{17.0} & \vg{75.9}{4.4} & \vg{75.5}{14.8} & \vg{93.4}{6.6} & \vg{94.4}{6.3} & \vg{83.1}{9.8} & \vg{59.1}{18.9} & \vg{47.1}{5.8} & \vg{53.1}{12.3} & \vg{19.1}{7.9} \\
\rowcolor{hlrow}
InternVL3.5-8B$^\dagger$
  & \vg{64.0}{22.3} & \vg{61.0}{5.6} & \vg{58.6}{16.4} & \vg{75.7}{21.6} & \vg{83.6}{3.8} & \vg{68.6}{14.0} & \vg{70.1}{7.1} & \vg{57.8}{7.5} & \vg{63.9}{7.2} & \vg{9.5}{4.7} \\
\bottomrule
\end{tabular}%
}
\end{table}

\paragraph{SFT gains transfer consistently across architectures and benchmarks.}
Table~\ref{tab:transfer} shows that \trainsuite{} improves performance on all five public MCQ benchmarks, both iNat2021 splits, and WorldVQA across all four evaluated models.
The strongest transfer appears in the 2B setting, where public-benchmark average rises by 14.5\,pp and iNat21 average by 16.2\,pp; the same pattern holds at larger scales and across architectures.
These results confirm that the broad category coverage of \trainsuite{} translates into consistent gains on both fine-grained recognition benchmarks and visual knowledge benchmarks, demonstrating its effectiveness as a general-purpose resource for improving visual world knowledge in MLLMs.

\section{Analysis}
\label{sec:analysis}

\subsection{Domain and Depth Diagnostics}
\label{sec:profiles}

\begin{figure}[t]
\centering
\includegraphics[width=\textwidth]{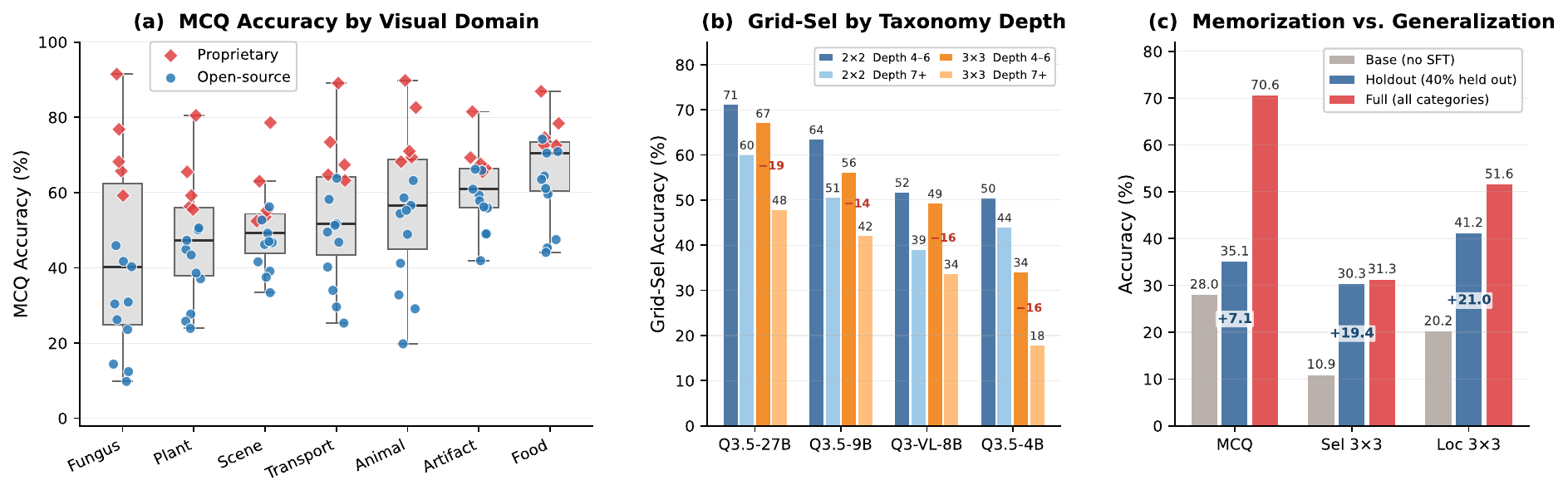}
\caption{Capability profiles vary substantially across visual domains, taxonomy depth, and category exposure.
\textbf{(a)}~MCQ accuracy by visual domain
(red~$\diamond$: proprietary; blue~$\circ$: open-source).
Cross-domain gaps exceed 33\,pp within individual models.
\textbf{(b)}~Grid-Sel accuracy by taxonomy depth for four open-source models
(Qwen3.5-27B/9B/4B, Qwen3-VL-8B) at 2$\times$2 and 3$\times$3 resolutions.
Performance degrades substantially at the deepest levels~(7+).
\textbf{(c)}~Memorization vs.\ generalization (Qwen3-VL-2B):
\textit{Base}~= instruct baseline (no SFT);
\textit{Holdout}~= fine-tuned excluding records with held-out ground-truth labels;
\textit{Full}~= fine-tuned on the complete SFT data.
Selection and localization improve on categories held out from SFT target labels, with smaller gains on MCQ.}
\label{fig:analysis_combined}
\end{figure}

\paragraph{Domain-level analysis reveals divergent weaknesses across model tiers.}
Figure~\ref{fig:analysis_combined}a shows that all models perform unevenly across domains, but the profile diverges between closed-source and open-source families.
The starkest example is fungus: all open-source models score lowest in this domain (e.g., Qwen3-VL-2B 9.8\%, Qwen3-VL-8B 30.4\%), yet Gemini~3.1~Pro achieves 91.5\%.
Within-model cross-domain gaps exceed 33\,pp for Qwen3-VL-8B and 29\,pp for Qwen3.5-27B, confirming that a single aggregate score can obscure important domain-specific weaknesses.

\paragraph{Performance degrades systematically with taxonomy depth.}
Figure~\ref{fig:analysis_combined}b shows that representative open-source models consistently degrade from intermediate depths (4--6) to the deepest levels (7+), especially in the 3$\times$3 setting: Qwen3.5-27B drops from 67.1\% to 47.9\% on Grid-Sel 3$\times$3, a decline of 19.2\,pp.
Together, these domain and depth analyses show that aggregate scores mask systematic failure modes visible only under taxonomy-stratified evaluation.

\subsection{Transfer under Target-Label Holdout}
\label{sec:holdout}

Using Qwen3-VL-2B, we randomly hold out 40\% of benchmark categories and remove all SFT records whose ground-truth labels belong to these categories. On benchmark instances with held-out ground-truth labels, we compare the instruct baseline (no SFT), the holdout model (filtered SFT data), and the full model (complete SFT data).
Experimental details are provided in Appendix~\ref{sec:holdout_experiment}.

\paragraph{Contrastive selection and localization show stronger transfer than closed-set recognition.}
Figure~\ref{fig:analysis_combined}c shows that the holdout model substantially outperforms the instruct baseline on Grid-Sel 3$\times$3 (10.9 $\rightarrow$ 30.3) and Grid-Loc 3$\times$3 (20.2 $\rightarrow$ 41.2), although these categories do not appear as ground-truth labels in its SFT data.
On MCQ, by contrast, the holdout model improves only modestly ($+$7.1\,pp) while the full model gains $+$42.6\,pp, suggesting greater sensitivity of MCQ to the target-label holdout setting.
These results support transfer beyond categories directly supervised as SFT targets, with larger gains on selection and localization than on MCQ.

\section{Conclusion}
\label{sec:conclusion}

We presented \bench{}, a taxonomy-grounded multi-task diagnostic benchmark evaluating 15{,}479 ground-truth categories within a taxonomy of 21{,}291 leaf-node classes, together with \trainsuite{}, a 50K+-category training resource. Systematic evaluation across closed-source, open-source, and fine-tuned models shows that visual world knowledge is inherently multi-dimensional: strong closed-set recognition does not imply strong contrastive selection, precise localization, or open-world naming, and these dissociations persist across model families. Under targeted supervision, fine-tuning on \trainsuite{} yields substantial gains on \bench{} and transfers consistently to public fine-grained recognition and visual knowledge benchmarks. Under target-label holdout, contrastive selection and localization show larger gains over the instruct baseline than closed-set recognition, supporting transfer to categories not directly supervised as SFT targets. Together, these findings suggest that visual world knowledge should be evaluated and improved as a multi-dimensional capability rather than reduced to a single recognition score.

\begingroup
\small
\bibliographystyle{unsrtnat}
\bibliography{references}
\endgroup

\clearpage
\appendix
{\LARGE\bfseries Appendix\par}
\vspace{1em}
\setcounter{section}{0}
\renewcommand{\thesection}{\Alph{section}}

\definecolor{okgreen}{HTML}{2E7D32}
\definecolor{errred}{HTML}{C62828}
\definecolor{lgray}{HTML}{888888}

\section{Limitations}
\label{sec:appendix_limitations}

Our SFT experiments focus on the Qwen3-VL and InternVL3.5 backbone families, so findings may not fully generalize to other architectures.
Open-QA uses EM+LLM Judge; stronger alias-aware and hierarchy-aware metrics remain future work.
Grid-Sel and Grid-Loc use controlled 2$\times$2 / 3$\times$3 layouts that are diagnostically useful but do not fully match natural multi-object scenes, and Grid-Loc may additionally reflect sensitivity to normalized box formatting.
The benchmark spans eight visual domains, but specialized areas such as medical imaging or satellite imagery remain under-covered.

\section{Broader impacts}
\label{sec:broader_impacts}
This work provides a diagnostic benchmark and training resource for fine-grained visual recognition. It can support research on species and object recognition, reveal gaps in models' visual world knowledge, and help researchers compare capabilities across complementary tasks.

Potential negative impacts arise from misidentification and over-reliance on benchmark performance. Incorrect species or object predictions could mislead users, particularly when applied to decisions such as whether a plant or fungus is safe to consume. The benchmark's uneven domain coverage and controlled grid layouts also limit how well aggregate scores represent performance in natural settings.

The benchmark is intended for research evaluation. Its scores do not establish safety or reliability for field species identification, food-safety judgments, or other consequential applications. Such uses require application-specific validation and appropriate expert review; model predictions should not serve as the sole basis for these decisions.

\section{Benchmark Statistics}
\label{sec:bench_stats_appendix}

\begin{table}[h]
\centering
\caption{Statistics of \bench{}.
\textit{\#Cls}: unique canonical ground-truth names in the released benchmark split.
The benchmark is built on a taxonomy of 21{,}291 leaf-node classes; the released evaluation subset covers 15{,}479 unique ground-truth categories, while nearly all taxonomy classes appear as distractors.
Grid-Loc is evaluated at IoU threshold $\tau{=}0.5$ following the visual grounding standard.
Grid tasks are evaluated at two spatial resolutions (2$\times$2 and 3$\times$3) to probe fine-grained discrimination at different spatial granularities.
}
\label{tab:bench_stats}
\small
\begin{tabular}{@{}llrrl@{}}
\toprule
\textbf{Task} & \textbf{Format} & \textbf{\#Samples} & \textbf{\#Cls} & \textbf{Metric} \\
\midrule
MCQ & 20-option MC & 14{,}717 & 11{,}466 & Accuracy \\
Grid-Sel 2$\times$2 & Select cell & 4{,}085 & 3{,}726 & Accuracy \\
Grid-Sel 3$\times$3 & Select cell & 4{,}125 & 3{,}785 & Accuracy \\
Grid-Loc 2$\times$2 & Output bbox & 3{,}959 & 3{,}613 & IoU@0.5 \\
Grid-Loc 3$\times$3 & Output bbox & 3{,}446 & 3{,}204 & IoU@0.5 \\
Open-QA & Free-form text & 2{,}209 & 2{,}187 & EM + LLM Judge \\
\midrule
\textbf{Total} & & \textbf{32{,}541} & \textbf{15{,}479 unique} & \\
\bottomrule
\end{tabular}
\end{table}

\paragraph{Release pool and retention.}
The released benchmark is assembled from an intermediate task-specific pool that already satisfies the main verification and release-stage constraints.
For MCQ, this pool contains 22{,}356 samples before the final image-quality filter; the corresponding counts are 5{,}559 for Grid-Sel 2$\times$2, 5{,}512 for Grid-Sel 3$\times$3, 5{,}383 for Grid-Loc 2$\times$2, 4{,}606 for Grid-Loc 3$\times$3, and 6{,}169 for Open-QA.
After automated image-quality filtering that removes low-quality images (too dark, too small, blurry, low saturation, etc.) and additional quality-score-based curation for Grid and Open-QA tasks, the released benchmark retains 65.8\% of MCQ, 73.5\% of Grid-Sel 2$\times$2, 74.8\% of Grid-Sel 3$\times$3, 73.5\% of Grid-Loc 2$\times$2, 74.8\% of Grid-Loc 3$\times$3, and 35.8\% of Open-QA, yielding a total of 32{,}541 evaluation instances.

\begin{table}[h]
\centering
\caption{Per-domain sample counts in the released benchmark split for MCQ.
Grid tasks and Open-QA follow the same 8-domain taxonomy; per-domain breakdowns are omitted for brevity.
The released split is intentionally not domain-balanced, so sparse domains such as Scene and especially Other should not be over-interpreted as standalone evaluation subsets.}
\label{tab:bench_domain_stats}
\small
\begin{tabular}{@{}lr@{}}
\toprule
\textbf{Domain} & \textbf{MCQ} \\
\midrule
Animal & 4{,}011 \\
Artifact & 2{,}535 \\
Food & 2{,}411 \\
Plant & 2{,}030 \\
Fungus & 1{,}696 \\
Transport & 1{,}664 \\
Scene & 368 \\
Other & 2 \\
\midrule
\textbf{Total} & \textbf{14{,}717} \\
\bottomrule
\end{tabular}
\end{table}

\begin{table}[h]
\centering
\caption{MCQ option composition in the released benchmark split. Percentages use construction-time option types over all slots, including the ground-truth option; these types are omitted from public \texttt{mcq.jsonl}.}
\label{tab:mcq_option_composition}
\small
\begin{tabular}{@{}lrr@{}}
\toprule
\textbf{Option type} & \textbf{Count} & \textbf{Ratio} \\
\midrule
Sibling & 111{,}866 & 38.0\% \\
Random & 108{,}327 & 36.8\% \\
Near & 59{,}430 & 20.2\% \\
GT & 14{,}717 & 5.0\% \\
\bottomrule
\end{tabular}
\end{table}

\paragraph{GT-category filtering check.}
We additionally verify that the released MCQ split does not contain low-confidence Bamboo leaf nodes from the initial taxonomy audit.
We compare the 6{,}948 low-confidence categories identified in the initial taxonomy audit against the 11{,}466 unique construction-time ground-truth IDs corresponding to the released MCQ split (14{,}717 samples in total).
The intersection is zero, confirming that the released MCQ ground-truth labels exclude all audited Bamboo leaf categories with \texttt{name\_match} $< 0.7$. The simplified public files do not expose the construction-time \texttt{gt\_id} field.
This verification applies only to ground-truth labels and does not constrain distractor options.

\section{Taxonomy Construction}
\label{sec:taxonomy_details}

The benchmark taxonomy contains 21{,}291 leaf-node classes across eight visual domains: animal, plant, fungus, artifact, transport, food, scene, and other.
The taxonomy is derived from the Bamboo category tree~\cite{bamboo} and further expanded with supplementary public sources to improve category coverage across domains.
After normalization, deduplication, filtering, and manual audit over both the source category set and the associated image pool, only leaf-level categories with visually identifiable image evidence are retained as benchmark ground-truth labels.
\trainsuite{} further expands category coverage beyond the benchmark taxonomy by incorporating additional public fine-grained datasets, including iNaturalist~2021.

\paragraph{Normalization and filtering.}
Category names are normalized to a canonical English label with optional aliases.
Near-duplicate labels, spelling variants, and parent-child duplicates are merged when they refer to the same visual category.
Categories are removed when they are non-visual, overly broad for fine-grained recognition, visually unverifiable from available images, or incompatible with the benchmark's taxonomy structure.

\begin{table}[h]
\centering
\caption{Per-domain taxonomy statistics.
\textit{\#Leaf classes}: total leaf-node categories in the taxonomy.
\textit{\#GT classes}: unique ground-truth categories appearing in the released benchmark across all tasks.
Nearly all 21{,}291 leaf classes additionally appear as distractors in MCQ or Grid tasks, even when they are not used as ground-truth labels.}
\label{tab:taxonomy_stats}
\small
\begin{tabular}{@{}lrrr@{}}
\toprule
\textbf{Domain} & \textbf{\#Leaf classes} & \textbf{\#GT classes} & \textbf{Max depth} \\
\midrule
Animal & 6{,}620 & 5{,}323 & 13 \\
Plant & 2{,}198 & 1{,}786 & 8 \\
Fungus & 1{,}690 & 1{,}492 & 6 \\
Artifact & 4{,}819 & 2{,}753 & 14 \\
Transport & 2{,}597 & 1{,}859 & 8 \\
Food & 2{,}373 & 1{,}853 & 9 \\
Scene & 433 & 359 & 7 \\
Other & 561 & 54 & 10 \\
\midrule
\textbf{Total} & \textbf{21{,}291} & \textbf{15{,}479} & \textbf{14} \\
\bottomrule
\end{tabular}
\end{table}

\section{Task Construction}
\label{sec:task_construction_appendix}

All four tasks are derived from the same taxonomy-grounded candidate pool. Public files provide task inputs, answer annotations, and domain/depth metadata, but omit construction-time category IDs and parent-category fields.
This shared construction makes the task suite diagnostically comparable rather than a collection of unrelated benchmarks.

\paragraph{MCQ construction.}
Each released MCQ sample contains an image path, 20 candidate category names as strings in \texttt{options}, and a zero-based correct-answer index in \texttt{gt\_index}, together with \texttt{qid} and domain/depth metadata.
The sibling, near, and random distinctions describe distractor construction; the simplified public files do not include per-option category IDs or type labels.
The benchmark is built by reorganizing and normalizing previously quality-checked bench pools rather than sampling directly from the full raw image archive.
For each ground-truth class, we retain at most two benchmark samples, prioritizing wider class coverage first and then applying the release-balancing rules described below.

\paragraph{Grid-Sel construction.}
Each Grid-Sel sample is a composite image formed by tiling one target image together with distractor images from taxonomically neighboring categories.
We construct both 2$\times$2 and 3$\times$3 grids.
Distractors are drawn with a sibling-first policy: we first sample direct siblings of the target from the taxonomy, and only expand to tree distance 2 when the sibling pool is insufficient.
Public Grid-Sel records provide the target category, prompt, and cell answer, but omit per-grid sibling counts and taxonomy-distance diagnostics.
Within a single grid, all source images are distinct, while exact image-set duplicates across grids are disallowed by a grid-level deduplication key.

\paragraph{Grid-Loc construction.}
Grid-Loc shares the same composite images as Grid-Sel but changes the output space from discrete cell prediction to precise bounding-box localization.
The target bounding box is not the full target cell.
Instead, it is the target object's bounding box after propagating the original object annotation through the crop/pad transform used to place the image into its grid cell, and then mapping that box into the coordinates of the full composite grid image.
The released benchmark stores this target box as \texttt{bbox\_norm}, with coordinates normalized to $[0,1000]$ relative to the full grid image width and height.

\paragraph{Open-QA construction.}
Open-QA is derived from high-quality MCQ upstream samples rather than created independently from the raw data pool.
Public Open-QA records provide image paths, questions, alias sets, and ancestor metadata; they omit the upstream MCQ linkage field \texttt{source\_qid}.
These metadata support exact matching, synonym handling, and future hierarchy-aware analysis, although the main paper reports only EM+LLM Judge.

\paragraph{Prompt design.}
The benchmark uses direct task-specific user prompts with constrained answer formats.
MCQ asks the model to return a single option index from 20 candidates.
Grid-Sel asks for a single cell number.
Grid-Loc requires exactly one JSON-style array \texttt{[x1, y1, x2, y2]} normalized to $[0,1000]$.
Open-QA asks the model to identify the object as specifically as possible and place the final answer on a dedicated \texttt{Answer:} line.
These prompts are instantiated deterministically from the benchmark metadata and are described further in Section~\ref{sec:evaluation_details}.

\begin{figure}[h]
\centering
\begin{minipage}[t]{0.48\textwidth}
\centering
\includegraphics[width=2.5cm,height=2.5cm,keepaspectratio]{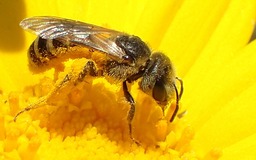}\\[1pt]
{\small\textbf{(a) MCQ \textbar\ Animal}}\\[1pt]
{\scriptsize \textbf{Q:} What species of insect is shown? \textit{(20 options)}}\\[1pt]
{\scriptsize \textbf{GT:} \textcolor{okgreen}{\textbf{halictus ligatus}}}\\[0pt]
{\scriptsize \textcolor{lgray}{lyssomanes viridis \textit{(near)}}}\\[0pt]
{\scriptsize \textcolor{lgray}{eratigena duellica \textit{(near)},\; ... +17 more}}
\end{minipage}
\hfill
\begin{minipage}[t]{0.48\textwidth}
\centering
\includegraphics[width=2.5cm,height=2.5cm,keepaspectratio]{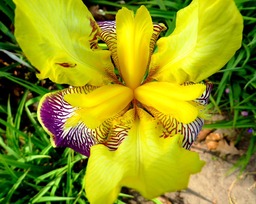}\\[1pt]
{\small\textbf{(b) MCQ \textbar\ Plant}}\\[1pt]
{\scriptsize \textbf{Q:} What species of plant is shown? \textit{(20 options)}}\\[1pt]
{\scriptsize \textbf{GT:} \textcolor{okgreen}{\textbf{iris (plant)}}}\\[0pt]
{\scriptsize \textcolor{lgray}{freesia \textit{(near)}}}\\[0pt]
{\scriptsize \textcolor{lgray}{gladiolus \textit{(sibling)},\; ... +17 more}}
\end{minipage}
\vspace{4pt}
\hrule
\vspace{4pt}
\begin{minipage}[t]{0.48\textwidth}
\centering
\includegraphics[width=2.5cm,height=2.5cm,keepaspectratio]{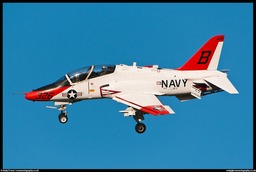}\\[1pt]
{\small\textbf{(c) MCQ \textbar\ Transport}}\\[1pt]
{\scriptsize \textbf{Q:} What aircraft model is shown? \textit{(20 options)}}\\[1pt]
{\scriptsize \textbf{GT:} \textcolor{okgreen}{\textbf{T-45 Goshawk}}}\\[0pt]
{\scriptsize \textcolor{lgray}{Lockheed Martin X-35 \textit{(near)}}}\\[0pt]
{\scriptsize \textcolor{lgray}{C-17A Globemaster III \textit{(near)},\; ... +17 more}}
\end{minipage}
\hfill
\begin{minipage}[t]{0.48\textwidth}
\centering
\includegraphics[width=2.5cm,height=2.5cm,keepaspectratio]{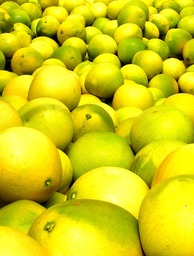}\\[1pt]
{\small\textbf{(d) MCQ \textbar\ Plant}}\\[1pt]
{\scriptsize \textbf{Q:} What kind of fruit is shown? \textit{(20 options)}}\\[1pt]
{\scriptsize \textbf{GT:} \textcolor{okgreen}{\textbf{pomelo}}}\\[0pt]
{\scriptsize \textcolor{lgray}{grapefruit \textit{(sibling)}}}\\[0pt]
{\scriptsize \textcolor{lgray}{raspberry \textit{(random)},\; ... +17 more}}
\end{minipage}
\vspace{4pt}
\hrule
\vspace{4pt}
\begin{minipage}[t]{0.48\textwidth}
\centering
\includegraphics[width=2.5cm,height=2.5cm,keepaspectratio]{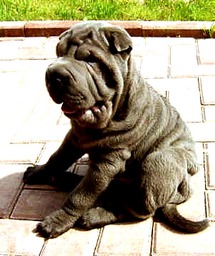}\\[1pt]
{\small\textbf{(e) Open-QA \textbar\ Animal}}\\[1pt]
{\scriptsize \textbf{Q:} What breed of dog is shown? \textit{(open-ended)}}\\[1pt]
{\scriptsize \textbf{GT:} \textcolor{okgreen}{\textbf{shar pei}}}\\[0pt]
{\scriptsize No candidate options provided;}\\[0pt]
{\scriptsize the model must generate the name.}
\end{minipage}
\hfill
\begin{minipage}[t]{0.48\textwidth}
\centering
\includegraphics[width=2.5cm,height=2.5cm,keepaspectratio]{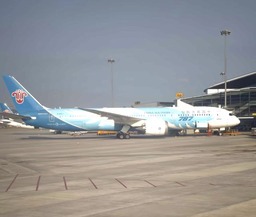}\\[1pt]
{\small\textbf{(f) Open-QA \textbar\ Transport}}\\[1pt]
{\scriptsize \textbf{Q:} What aircraft model is shown? \textit{(open-ended)}}\\[1pt]
{\scriptsize \textbf{GT:} \textcolor{okgreen}{\textbf{Boeing 787 Dreamliner}}}\\[0pt]
{\scriptsize No candidate options provided;}\\[0pt]
{\scriptsize the model must generate the name.}
\end{minipage}
\caption{MCQ and Open-QA task examples. (a)--(d) MCQ: closed-set recognition with 20 taxonomy-aware options including sibling, near, and random distractors drawn from the taxonomy hierarchy. (e)--(f) Open-QA: the model must generate the precise fine-grained category name without any candidate set.}
\label{fig:case_mcq_oqa}
\end{figure}

\begin{figure}[h]
\centering
\begin{minipage}[t]{0.48\textwidth}
\centering
\includegraphics[width=3.6cm]{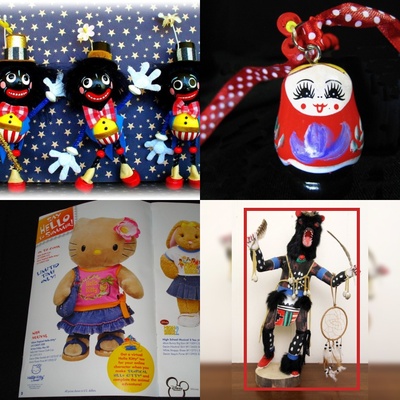}\\[2pt]
{\small\textbf{(a) Grid-Sel \textbar\ Artifact}}\\[1pt]
{\scriptsize \textbf{Target:} \textcolor{okgreen}{\textbf{kachina}} (Cell 4)}\\[1pt]
{\scriptsize \textcolor{lgray}{Cell 1: golliwog}}\\[0pt]
{\scriptsize \textcolor{lgray}{Cell 2: matryoshka doll}}\\[0pt]
{\scriptsize \textcolor{lgray}{Cell 3: Hello Kitty}}\\[1pt]
{\scriptsize \textcolor{errred}{\rule{6pt}{2pt}} GT bbox}
\end{minipage}
\hfill
\begin{minipage}[t]{0.48\textwidth}
\centering
\includegraphics[width=3.6cm]{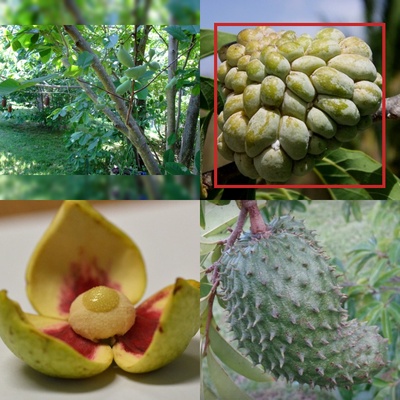}\\[2pt]
{\small\textbf{(b) Grid-Sel \textbar\ Plant}}\\[1pt]
{\scriptsize \textbf{Target:} \textcolor{okgreen}{\textbf{sweetsop}} (Cell 2)}\\[1pt]
{\scriptsize \textcolor{lgray}{Cell 1: asimina triloba}}\\[0pt]
{\scriptsize \textcolor{lgray}{Cell 3: annona glabra}}\\[0pt]
{\scriptsize \textcolor{lgray}{Cell 4: soursop}}\\[1pt]
{\scriptsize \textcolor{errred}{\rule{6pt}{2pt}} GT bbox}
\end{minipage}
\vspace{6pt}
\hrule
\vspace{6pt}
\begin{minipage}[t]{0.48\textwidth}
\centering
\includegraphics[width=3.6cm]{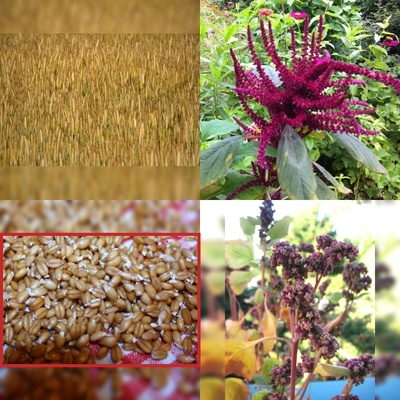}\\[2pt]
{\small\textbf{(c) Grid-Loc \textbar\ Food}}\\[1pt]
{\scriptsize \textbf{Target:} \textcolor{okgreen}{\textbf{wheat berry}} (Cell 3)}\\[1pt]
{\scriptsize \textcolor{lgray}{Cell 1: rye \quad Cell 2: amaranth}}\\[0pt]
{\scriptsize \textcolor{lgray}{Cell 4: Chenopodium quinoa}}\\[1pt]
{\scriptsize \textcolor{errred}{\rule{6pt}{2pt}} GT bbox}
\end{minipage}
\hfill
\begin{minipage}[t]{0.48\textwidth}
\centering
\includegraphics[width=3.6cm]{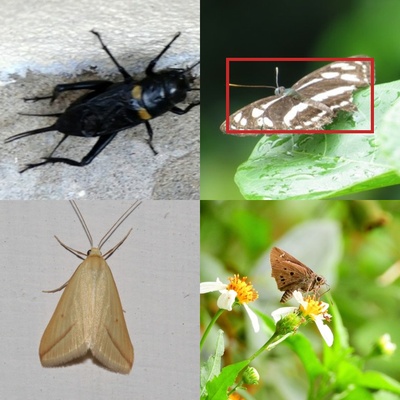}\\[2pt]
{\small\textbf{(d) Grid-Loc \textbar\ Animal}}\\[1pt]
{\scriptsize \textbf{Target:} \textcolor{okgreen}{\textbf{phaedyma columella}} (Cell 2)}\\[1pt]
{\scriptsize \textcolor{lgray}{Cell 1: gryllus bimaculatus}}\\[0pt]
{\scriptsize \textcolor{lgray}{Cell 3: rhodometra sacraria}}\\[0pt]
{\scriptsize \textcolor{lgray}{Cell 4: suastus gremius}}\\[1pt]
{\scriptsize \textcolor{errred}{\rule{6pt}{2pt}} GT bbox}
\end{minipage}
\caption{Grid task examples (2$\times$2). (a)--(b) Grid-Sel: the model selects which cell contains the target category from a composite grid of taxonomically similar images. (c)--(d) Grid-Loc: the model outputs a bounding box to precisely localize the target object within the grid. \textcolor{errred}{Red} boxes indicate the ground-truth bounding box. All distractors are sibling or near categories from the taxonomy.}
\label{fig:case_grid}
\end{figure}

\section{Quality Control and Curation}
\label{sec:qc_details}

\paragraph{Annotation verification.}
Each candidate sample is first evaluated by three frontier closed-source models (Gemini~3~Pro, GPT-5, GPT-4o).
These QC assistants are distinct from the evaluated model versions / endpoints in Section~\ref{sec:experiments}, although some belong to the same product family (e.g., GPT-5 for QC versus GPT-5.2 / GPT-5.4 for evaluation).
Samples for which at least two of the three models produce an incorrect answer are flagged for human review.
These model outputs are used only for triage: flagged samples are not removed automatically, and final keep/remove decisions are made under predefined human criteria (label correctness, visual identifiability, ambiguity, and leakage checks).
Human annotators then assign each flagged sample to one of three outcomes: labeling error, ambiguous sample, or valid reviewed case.
Labeling errors and ambiguous samples are removed, while valid reviewed cases are retained.

\subsection{Audit of Non-flagged Samples}
\label{sec:qc_audit}

\paragraph{Random audit of non-flagged samples.}
To test whether model-assisted triage systematically misses problematic samples that most assisting models answer correctly, we additionally conduct a random human audit on non-flagged samples from each task.
The audit is performed under the same criteria as the main verification pass and is used only to assess the robustness of the QC pipeline, not to redefine the benchmark.
Table~\ref{tab:qc_audit} reports the audit protocol and issue rates.

\paragraph{Grid-specific checks.}
For Grid-Sel and Grid-Loc, verification is performed at the per-cell level to ensure that each grid cell matches its assigned category.
Samples are removed if the target object occupies less than 1\% of the image area, if the grid composition is visually corrupted, or if the target category is not visually distinguishable from the distractor cells.
Grid-Loc uses the same composite-image pool as Grid-Sel after alignment, but applies the IoU@0.5 evaluation threshold.

\paragraph{Open-QA checks.}
Open-QA candidates are filtered in two stages.
First, samples whose ground-truth categories are not visually identifiable are removed.
Second, a frontier model performs visual quality inspection for image quality, text leakage, identifiability, and ground-truth correctness; only samples receiving a KEEP verdict are retained.

\begin{table}[h]
\centering
\caption{Quality-control statistics. The \textit{Release pool} column reports the count after the task-specific intermediate filtering stage (release balancing for MCQ and Grid-Sel, visual QC for Grid-Loc and Open-QA) but before the final image-quality filter. The \textit{Final} column reports the count after the additional automated image-quality filter.}
\label{tab:qc_stats}
\small
\begin{tabular}{@{}lrrrr@{}}
\toprule
\textbf{Task} & \textbf{Candidate pool} & \textbf{Release pool} & \textbf{Final} & \textbf{Retention} \\
\midrule
MCQ & 47{,}697 & 22{,}356 & 14{,}717 & 30.9\% \\
Grid-Sel 2$\times$2 & 19{,}131 & 5{,}559 & 4{,}085 & 21.4\% \\
Grid-Sel 3$\times$3 & 18{,}064 & 5{,}512 & 4{,}125 & 22.8\% \\
Grid-Loc 2$\times$2 & 18{,}681 & 5{,}383 & 3{,}959 & 21.2\% \\
Grid-Loc 3$\times$3 & 15{,}335 & 4{,}606 & 3{,}446 & 22.5\% \\
Open-QA & 10{,}000 & 6{,}169 & 2{,}209 & 22.1\% \\
\midrule
\textbf{Total} & \textbf{128{,}908} & & \textbf{32{,}541} & \textbf{25.2\%} \\
\bottomrule
\end{tabular}
\end{table}

\begin{table}[h]
\centering
\caption{Random audit of non-flagged samples.
200 samples per task are drawn uniformly at random from the non-flagged pool (seed\,=\,42) and reviewed under the same criteria as the main verification pass.
\textit{Issue}: any sample requiring correction or removal because of label error, ambiguity, non-identifiability, leakage, or corrupted composition.}
\label{tab:qc_audit}
\small
\begin{tabular}{@{}lrrrr@{}}
\toprule
\textbf{Task} & \textbf{Audited} & \textbf{Issue} & \textbf{Issue rate} & \textbf{Main issue types} \\
\midrule
MCQ & 200 & 6 & 3.0\% & ambiguity \\
Grid-Sel & 200 & 8 & 4.0\% & cell mismatch \\
Grid-Loc & 200 & 8 & 4.0\% & box misalignment \\
Open-QA & 200 & 5 & 2.5\% & alias ambiguity \\
\midrule
\textbf{Total} & \textbf{800} & \textbf{27} & \textbf{3.4\%} & \\
\bottomrule
\end{tabular}
\end{table}

\paragraph{Release construction and auxiliary metadata.}
Response-pattern labels are derived from frontier-model outputs and used only as QC / audit metadata for identifying potentially problematic samples for additional review. They are not release-construction criteria, stand-alone validity criteria, or the basis for defining benchmark difficulty.

\textit{MCQ release construction.}
For MCQ, we begin the final release stage from an intermediate pool of 22{,}356 samples that already satisfies the main verification and release-balancing constraints.
Importantly, samples are \emph{not} removed simply because reference models answer them correctly; rather, model disagreement and repeated failure patterns are used only to trigger additional human verification for potentially problematic cases.
We then apply a final automated image-quality filter that removes visually unreliable images (too dark, too small, blurry, low saturation), yielding 14{,}717 released MCQ samples.

\textit{Grid-Sel release construction} follows the same principle: final release balancing is carried out over verified samples while jointly considering distractor structure, domain coverage, and visual quality, followed by five-dimension visual QC (target-cell quality, text leakage, distinguishability, ground-truth correctness, overall verdict) using dual frontier models (Gemini~3~Pro primary, GPT-5 fallback), retaining only KEEP samples.

\textit{Grid-Loc} is evaluated on the same composite images as Grid-Sel after alignment; the IoU@0.5 threshold follows the standard used in visual grounding benchmarks.

\textit{Open-QA.}
Three frontier models (Gemini~3~Pro, GPT-5, GPT-4o) independently attempt each question via exact match.
These outputs define three auxiliary response-pattern buckets: \texttt{full\_consensus} (3/3 correct), \texttt{partial\_consensus} (1--2/3 correct), and \texttt{zero\_consensus} (0/3 correct).
These buckets are retained only as auxiliary audit metadata; the released set is assembled from verified samples under domain filtering (removing categories with non-visual ground truths), the same five-dimension visual QC as Grid, and automated image-quality filtering, yielding 2{,}209 samples.

\subsection{Sanity Check on Model-Assisted QC}
\label{sec:model_dependence_sanity}

By design, the frontier models used for annotation triage and QC (Gemini~3~Pro, GPT-5, GPT-4o) are version-distinct from the evaluated model endpoints in Section~\ref{sec:experiments}, eliminating direct circular dependence between curation and evaluation at the deployed-model level.
These model outputs are used to identify samples that require additional verification; they are not used as automatic deletion rules for samples answered correctly by the assisting models.
As additional evidence that the main findings are not artifacts of any particular assisting model family, Table~\ref{tab:model_dependence_sanity} confirms that the cross-task capability dissociations (large MCQ--Open and MCQ--Loc gaps) persist across diverse closed-source, open-source, and SFT model families.

\begin{table}[h]
\centering
\caption{Sanity check: cross-task capability dissociations across model families.
\textit{MCQ--Open}: difference between closed-set MCQ accuracy and Open-QA accuracy.
\textit{MCQ--Loc}: difference between MCQ accuracy and Grid-Loc 3$\times$3 accuracy.
Large gaps persist across all families, confirming that the dissociations are general rather than artifacts of any assisting model.}
\label{tab:model_dependence_sanity}
\small
\setlength{\tabcolsep}{4pt}
\begin{tabular}{@{}llrrrr@{}}
\toprule
\textbf{Model} & \textbf{Family} & \textbf{MCQ} & \textbf{Open} & \textbf{MCQ--Open} & \textbf{MCQ--Loc} \\
\midrule
Gemini 3.1 Pro & Closed-source & 86.0 & 60.6 & 25.4 & 2.7 \\
Qwen3.5-27B & Open-source & 59.1 & 28.7 & 30.4 & 2.9 \\
Gemma-4-31B & Open-source & 60.5 & 27.2 & 33.3 & 24.4 \\
InternVL3.5-8B & Open-source & 35.5 & 10.3 & 25.2 & 28.1 \\
Qwen3-VL-8B$^\dagger$ & SFT & 80.0 & 33.9 & 46.1 & 15.7 \\
\bottomrule
\end{tabular}
\end{table}

\section{Training Details}
\label{sec:training_details}

All SFT models are trained with LoRA in LLaMA-Factory~\cite{zheng2024llamafactory} for 3 epochs using AdamW, learning rate $10^{-4}$, and an effective global batch size of 1024.
Unless otherwise stated, we use LoRA~\cite{hu2022lora} with rank 8 and target all trainable modules.
\trainsuite{} is anchored by large-scale closed-set MCQ supervision, with a reasoning-augmented MCQ subset and smaller grounding and open-ended QA subsets providing complementary signals.

\begin{table}[h]
\centering
\caption{\trainsuite{} composition. The resource is dominated by MCQ supervision; reasoning-augmented MCQ is a subset of the MCQ records rather than an additional disjoint pool.}
\label{tab:training_suite_stats}
\small
\begin{tabular}{@{}lrrl@{}}
\toprule
\textbf{Format} & \textbf{\#Records} & \textbf{\#Classes} & \textbf{Primary signal} \\
\midrule
MCQ & {$\sim$}6.9M & 48{,}170 & Closed-set recognition \\
Reasoning-augmented MCQ & {$\sim$}0.6M & 25{,}376 & Fine-grained visual rationale \\
Bounding-box grounding & {$\sim$}0.4M & 31{,}122 & Spatial grounding \\
Open-ended QA & {$\sim$}0.4M & 31{,}033 & Category-name generation \\
\midrule
\textbf{Total} & {$\sim$}7.7M & 50K+ & Multi-format fine-grained supervision \\
\bottomrule
\end{tabular}
\end{table}

For each SFT model, we inspect late-stage checkpoints to confirm convergence and report one representative converged checkpoint in the main paper.
Table~\ref{tab:sft_compute} summarizes the compute used by the four main SFT runs. Compute usage for the holdout run is not reported separately.
The 2B run used per-device batch size 64 and gradient accumulation 2 on 8 H200 GPUs; for the 4B and 8B runs, we adjusted the per-device batch size and accumulation settings to keep the same effective global batch size.

\begin{table}[h]
\centering
\caption{Compute resources for the main Qwen3-VL SFT, InternVL3.5 SFT runs. GPU-hours are computed as wall-clock hours multiplied by the number of GPUs.}
\label{tab:sft_compute}
\small
\begin{tabular}{@{}lrrrrr@{}}
\toprule
\textbf{Model} & \textbf{GPUs} & \textbf{Epochs} & \textbf{Global batch} & \textbf{Time} & \textbf{GPU-hours} \\
\midrule
Qwen3-VL-2B & 8$\times$H200 & 3 & 1024 & 32h & 256 \\
Qwen3-VL-4B & 16$\times$H200 & 3 & 1024 & 30h & 480 \\
Qwen3-VL-8B & 16$\times$H200 & 3 & 1024 & 45h & 720 \\
InternVL3.5-8B & 16$\times$H200 & 3 & 1024 & 41h & 656 \\
\bottomrule
\end{tabular}
\end{table}

\section{Evaluation Details}
\label{sec:evaluation_details}

Benchmark evaluation uses zero-shot prompting with greedy decoding.
MCQ and Grid-Sel are scored by accuracy, Grid-Loc uses IoU@0.5 following the standard threshold in visual grounding benchmarks~\cite{kazemzadeh2014referitgame,mao2016refcocog,flickr30k_entities}, and Open-QA uses EM+LLM Judge.
Grid-Sel and Grid-Loc are evaluated at both 2$\times$2 and 3$\times$3 resolutions; we report both settings as well as their within-task average.
For the reported average score, we take the arithmetic mean of MCQ, Sel Avg (mean of 2$\times$2 and 3$\times$3), Loc Avg (mean of 2$\times$2 and 3$\times$3), and Open-QA under EM+LLM Judge.

\paragraph{Prompt templates and output constraints.}
MCQ uses a prompt that presents the image together with 20 indexed candidate categories and instructs the model to output only the selected index.
Grid-Sel uses a prompt of the form: ``You are given a $N{\times}N$ grid image. Cells are numbered left-to-right, top-to-bottom. Which cell contains the target category: \texttt{\{class\_name\}}?'' and requires a single-number answer.
Grid-Loc uses a localization prompt that asks the model to return exactly one bounding box in the format \texttt{[x1, y1, x2, y2]}.
The coordinate convention is normalized to $[0,1000]$ relative to the full composite image, where $(0,0)$ is the top-left corner and $(1000,1000)$ is the bottom-right corner.
Open-QA asks the model to identify the object as specifically as possible, provide one brief evidence sentence, and then output the final prediction on a new line as \texttt{Answer: <category name>}.

\subsection{Prompt Templates}
\label{sec:prompt_templates}

Below we provide the exact prompt templates used for each task.
Placeholders are shown in \texttt{\{braces\}}.
All prompts are zero-shot with no system prompt.

\begin{tcolorbox}[
  colback=gray!5, colframe=gray!60, title={\textbf{MCQ Prompt}},
  fonttitle=\small, fontupper=\small, boxrule=0.4pt, arc=2pt
]
\ttfamily
Look at the object(s) in the image and choose the single best category from 1-\{N\}. Only output the number.\\[4pt]
1: \{option\_1\}\\
2: \{option\_2\}\\
\hspace*{0.5em}\vdots\\
20: \{option\_20\}
\end{tcolorbox}

\begin{tcolorbox}[
  colback=gray!5, colframe=gray!60, title={\textbf{Grid-Sel Prompt}},
  fonttitle=\small, fontupper=\small, boxrule=0.4pt, arc=2pt
]
\ttfamily
You are given a \{N\}x\{N\} grid image. Cells are numbered left-to-right, top-to-bottom as 1, 2, \ldots, \{N$\times$N\}. Which cell contains the target category: \{class\_name\}? Answer with a single number. Do not output any other text.
\end{tcolorbox}

\begin{tcolorbox}[
  colback=gray!5, colframe=gray!60, title={\textbf{Grid-Loc Prompt}},
  fonttitle=\small, fontupper=\small, boxrule=0.4pt, arc=2pt
]
\ttfamily
Locate the target object: \{class\_name\}.\\[4pt]
Output the bounding box as [x1, y1, x2, y2] where:\\
- (x1, y1) = top-left corner\\
- (x2, y2) = bottom-right corner\\
- Coordinates normalized to [0, 1000] (0=left/top edge, 1000=right/bottom edge)\\[4pt]
Example: [100, 200, 500, 800]\\[4pt]
Output only the four numbers in brackets.\\
Return exactly one bounding box for the target object.\\
Your entire reply must be exactly one single-line JSON array: [x1, y1, x2, y2]\\
Rules:\\
- Output exactly one JSON array with four numeric coordinates.\\
- Coordinates must be normalized to [0,1000].\\
- Do not output markdown, code fences, labels, keys, comments, explanations, prefixes, or suffixes.\\
- Do not output multiple boxes, alternatives, candidate answers, or reasoning.\\
- If uncertain, still output your single best box in the required format.
\end{tcolorbox}

\begin{tcolorbox}[
  colback=gray!5, colframe=gray!60, title={\textbf{Open-QA Prompt}},
  fonttitle=\small, fontupper=\small, boxrule=0.4pt, arc=2pt
]
\ttfamily

Identify the object in the image as specifically as possible.\\
\{question\}\\
First give one brief sentence describing the key visual evidence.\\
Then give the final answer on a new line in the format:\\
Answer: <category name>\\
Keep the evidence brief.\\
Do not add anything after the answer line.
\end{tcolorbox}

\vspace{0.5em}
\noindent
\textbf{Notes.}
(1)~MCQ options are numbered starting from 1; the model is expected to output a single integer.
(2)~Grid-Sel and Grid-Loc share the same composite grid image; only the output format differs (cell index vs.\ bounding box).
(3)~Grid-Loc coordinates follow the $[0,1000]$ normalization convention where $(0,0)$ is the top-left and $(1000,1000)$ is the bottom-right of the full composite image.
(4)~In Open-QA, \texttt{\{question\}} is the question provided with each sample and uses domain-aware phrasing (e.g., ``What kind of bird/flower/fungus\ldots'') to elicit fine-grained category-level responses rather than coarse object descriptions.

\paragraph{Answer parsing.}
For MCQ and Grid-Sel, a valid discrete answer index is extracted from the model response using fixed parsing rules and compared with the ground-truth index. Responses from which a valid index cannot be extracted are scored as incorrect.
For Grid-Loc, the evaluator searches the response for bracketed sequences of four numbers and uses the first matching sequence as the predicted box; surrounding explanatory text is permitted.
If no bracketed sequence is found, the evaluator falls back to the first four numeric values in the response.
Coordinates are clipped to $[0,1000]$, and reversed coordinate pairs are reordered along each axis.
The resulting box is compared against the stored normalized target box using IoU@0.5.
Responses from which four numeric values cannot be extracted are treated as parsing failures.
For Open-QA, the evaluator extracts the final answer line, normalizes the resulting string, applies exact matching against the canonical label and alias set, and forwards only non-exact cases to the LLM judge.

\paragraph{Decoding and failure handling.}
All reported results use greedy decoding with temperature $0$.
The benchmark prompts do not rely on separate task-specific system prompts.
For Grid-Loc, parsing failures among records not marked as request errors receive an IoU of zero and remain in the accuracy denominator.
Records marked as request errors are counted separately and excluded from the accuracy denominator.
Multiple-box responses are scored using the first extracted box and are additionally recorded in the multiple-box statistics.

\subsection{Open-QA Judging}
\label{sec:openqa_judge}

Open-QA answers are scored in three stages.
We first extract the model's final answer string, then match it against the canonical label and alias set.
For non-exact matches, a fixed frontier LLM judge (Gemini~2.5~Pro) is used to decide whether the prediction is semantically equivalent to the gold category, including capitalization differences, common-name/scientific-name variants, and accepted synonyms; this follows recent LLM-as-a-judge evaluation practice~\cite{geval,mtbench_judge}.
The judge is instructed not to accept predictions that only name a parent category or a visually similar sibling category.

\begin{tcolorbox}[
  colback=gray!5, colframe=gray!60, title={\textbf{Open-QA Judge Prompt (System)}},
  fonttitle=\small, fontupper=\small, boxrule=0.4pt, arc=2pt
]
\ttfamily
You are a taxonomy and fine-grained recognition expert.\\
Your task: determine whether a model's prediction refers to the SAME specific species/object as the ground truth.\\
Be strict --- only judge "correct" if they refer to the exact same species or object (synonyms, common name variants, and minor spelling differences are acceptable).\\
If the prediction is at a coarser level (e.g., genus instead of species, or "bird" instead of "American Robin"), judge "wrong".
\end{tcolorbox}

\begin{tcolorbox}[
  colback=gray!5, colframe=gray!60, title={\textbf{Open-QA Judge Prompt (User)}},
  fonttitle=\small, fontupper=\small, boxrule=0.4pt, arc=2pt
]
\ttfamily
Ground truth: \{gt\_name\}\\
Accepted aliases: \{aliases\}\\
Model's prediction: \{pred\_text\}\\[4pt]
Does the model's prediction refer to the same specific species/object as the ground truth?\\[4pt]
Output exactly one JSON object (no other text):\\
\{"judgment": "correct" or "wrong", "reason": "one sentence explanation"\}
\end{tcolorbox}

\noindent
Only samples that fail automated exact matching against the canonical label and alias set are forwarded to the LLM judge; exact matches are scored directly without an API call.

\begin{table}[h]
\centering
\caption{Open-QA judge validation on Doubao-Seed-2.0-pro predictions.
We sample 200 Gemini~2.5~Pro judge decisions (100 accept, 100 reject) on Open-QA predictions produced by Doubao-Seed-2.0-pro uniformly at random (seed\,=\,42) and independently re-evaluate them with a human annotator.
\textit{Precision}: fraction of judge-accepted predictions that are truly correct.
\textit{Recall}: fraction of truly correct predictions that the judge accepts. Cohen's $\kappa$ follows the standard inter-rater agreement coefficient~\cite{cohen1960coefficient}.}
\label{tab:openqa_judge_validation}
\small
\begin{tabular}{@{}lr@{}}
\toprule
\textbf{Metric} & \textbf{Value} \\
\midrule
Audited samples & 200 \\
Human--judge agreement & 94.0\% \\
Precision (judge accept) & 98.0\% \\
Recall (judge accept) & 90.7\% \\
F1 & 94.2\% \\
Cohen's $\kappa$ & 0.880 \\
False accept (judge accept, human incorrect) & 2 \\
False reject (judge reject, human correct) & 10 \\
\bottomrule
\end{tabular}
\end{table}

\section{Additional Analysis}
\label{sec:additional_analysis}

\subsection{Data Integrity}
\label{sec:data_integrity}

\paragraph{Decontamination procedure.}
Source datasets are filtered before training: (i) Cars-196 test images and (ii) iNaturalist~2021 validation images are removed wholesale from the training pool, since these images appear (or may appear) in standard public evaluation splits.
Category overlap, by contrast, is intrinsic to the benchmark design: \trainsuite{} and \bench{} are intended to cover the same fine-grained recognition space, so they are not category-disjoint.
We therefore treat image-level overlap as the relevant leakage criterion, while the holdout experiment (Appendix~\ref{sec:holdout_experiment}) probes transfer to categories excluded from SFT target labels.

\section{Holdout Experiment Details}
\label{sec:holdout_experiment}

\subsection{Experiment Design}
\label{sec:holdout_design}

We investigate whether fine-grained SFT improves performance on categories that are not directly supervised as SFT targets.
For this target-label holdout experiment, we randomly hold out 40\% of benchmark categories (seed\,=\,42) and remove \emph{all} SFT training samples whose ground-truth label belongs to the held-out set.
We then compare three conditions on the subset of \bench{} instances whose ground-truth category falls within the held-out set:

\begin{enumerate}[leftmargin=*, itemsep=1pt]
    \item \textbf{Instruct baseline}: Qwen3-VL-2B without any fine-grained SFT.
    \item \textbf{Holdout model}: Qwen3-VL-2B fine-tuned on the filtered \trainsuite{} (records with held-out ground-truth labels removed).
    \item \textbf{Full model}: Qwen3-VL-2B fine-tuned on the complete \trainsuite{} (all categories included).
\end{enumerate}

\noindent
Gains over the instruct baseline measure improvement on categories absent from the SFT target labels. This split controls target-label supervision during SFT; it does not establish absence from pretraining or from other roles such as distractors in the SFT data. Full-versus-holdout comparisons also differ in training-data volume, so they do not isolate the effect of category-specific supervision.

\subsection{Training Data Filtering}
\label{sec:holdout_filtering}

Starting from \trainsuite{} ({$\sim$}7.69M records after the source-level filtering described in Appendix~\ref{sec:data_integrity}), we remove all records whose ground-truth label matches the held-out category set.
Table~\ref{tab:holdout_filtering} reports the per-format filtering statistics.

\begin{table}[h]
\centering
\caption{Training data filtering for the holdout experiment.
Records are removed when their ground-truth label belongs to the held-out category set.}
\label{tab:holdout_filtering}
\small
\begin{tabular}{@{}lrrrr@{}}
\toprule
\textbf{Format} & \textbf{Before} & \textbf{After} & \textbf{Removed} & \textbf{Removal\,\%} \\
\midrule
MCQ               & 6{,}872K & 5{,}197K & 1{,}675K & 24.4\% \\
Bounding-box       & 435K    & 305K    & 130K    & 30.0\% \\
Open-ended QA      & 385K    & 262K    & 123K    & 31.9\% \\
\midrule
\textbf{Total}    & \textbf{7{,}692K} & \textbf{5{,}764K} & \textbf{1{,}928K} & \textbf{25.1\%} \\
\bottomrule
\end{tabular}
\end{table}

A post-hoc audit confirms that no held-out category appears among the ground-truth labels of the filtered SFT data.
The overall removal rate (25.1\%) is lower than the holdout ratio (40\%) because a large proportion of training categories are not present in the benchmark and are therefore fully retained.

\subsection{Evaluation Subset}
\label{sec:holdout_eval_subset}

We evaluate on released-benchmark instances whose ground-truth labels fall into the held-out category set.
After intersecting the held-out category set with the released benchmark splits, the evaluation subsets contain 5{,}729 MCQ samples, 1{,}653 Grid-Sel\,3$\times$3 samples, and 1{,}377 Grid-Loc\,3$\times$3 samples.
All three tasks use the same curated benchmark splits and scoring metrics as the main results table (IoU@0.5 for Grid-Loc).

\subsection{Results}
\label{sec:holdout_results}

Table~\ref{tab:holdout_results} reports the three-condition comparison on the held-out evaluation subset.

\begin{table}[h]
\centering
\caption{Holdout experiment results on held-out categories.
All numbers are percentages evaluated on released-benchmark instances whose ground-truth labels belong to the held-out category set.
$\Delta_{\text{H-B}}$: holdout model gain over instruct baseline.
}
\label{tab:holdout_results}
\small
\begin{tabular}{@{}lccc@{}}
\toprule
\textbf{Condition} & \textbf{MCQ} & \textbf{Sel\,3$\times$3} & \textbf{Loc\,3$\times$3} \\
\midrule
Instruct baseline   & 28.0 & 10.9 & 20.2 \\
Holdout model       & 35.1 & 30.3 & 41.2 \\
Full model          & 70.6 & 31.3 & 51.6 \\
\midrule
$\Delta_{\text{H-B}}$ & $+$7.1 & $+$19.4 & $+$21.0 \\
\bottomrule
\end{tabular}
\end{table}

\paragraph{Transfer to categories held out from SFT target labels.}
The holdout model achieves substantial gains over the instruct baseline on Grid-Sel\,3$\times$3 ($+$19.4\,pp) and Grid-Loc\,3$\times$3 ($+$21.0\,pp), although these categories do not appear as ground-truth labels in its SFT data.
On Grid-Sel\,3$\times$3, the holdout model reaches 30.3\%, close to the full model's 31.3\%, showing a similar selection gain when these categories are excluded from SFT target labels.
On Grid-Loc, the full model retains a clear advantage (51.6\% vs.\ 41.2\%), while the holdout model still improves substantially over the instruct baseline.

\paragraph{Closed-set recognition shows smaller holdout gains.}
On MCQ, the holdout model improves only modestly over the baseline ($+$7.1\,pp), while the full model gains $+$42.6\,pp.
The larger full-versus-holdout gap on MCQ suggests greater sensitivity to the target-label holdout setting, although the comparison also changes the amount and composition of SFT data.

\paragraph{Sanity checks.}
The instruct baseline and the full model each show small gaps between retained-category and held-out-category evaluation subsets across all three tasks ($\leq$1.2\,pp). These checks suggest similar aggregate difficulty for these two models, but do not rule out other differences between the subsets or training conditions.

\paragraph{Summary.}
Under this target-label holdout setting, Grid-Sel and Grid-Loc show larger gains over the instruct baseline than MCQ, supporting transfer beyond categories directly supervised as SFT targets. These task-dependent gains reinforce the value of multi-task evaluation without establishing a strict separation between memorization and generalization.

\end{document}